\pdfoutput=1

\documentclass[11pt]{article}

\usepackage[final]{acl}

\usepackage{times}
\usepackage{latexsym}

\usepackage[T1]{fontenc}

\usepackage[utf8]{inputenc}

\usepackage{microtype}

\usepackage{inconsolata}

\usepackage{graphicx}

\usepackage{color, xcolor}
\usepackage{booktabs}
\usepackage{multicol}
\usepackage{multirow, makecell, caption}
\usepackage{colortbl}
\usepackage{tikz}
\usepackage{arydshln}
\usepackage{pgfplots}
\usepackage{amsmath}
\usepackage{amssymb}
\usepackage{pgfplots}
\pgfplotsset{compat=1.18}
\usepackage{tabularx}
\usepackage{enumerate}
\definecolor{darkgreen}{RGB}{0, 100, 0}

\makeatletter
\def\adl@drawiv#1#2#3{%
        \hskip.5\tabcolsep
        \xleaders#3{#2.5\@tempdimb #1{1}#2.5\@tempdimb}%
                #2\z@ plus1fil minus1fil\relax
        \hskip.5\tabcolsep}
\newcommand{\cdashlinelr}[1]{%
  \noalign{\vskip\aboverulesep
           \global\let\@dashdrawstore\adl@draw
           \global\let\adl@draw\adl@drawiv}
  \cdashline{#1}
  \noalign{\global\let\adl@draw\@dashdrawstore
           \vskip\belowrulesep}}
\makeatother

\title{Order Doesn't Matter, But Reasoning Does:\\ Training LLMs with Order-Centric Augmentation}

\author{
    Qianxi He\textsuperscript{1}, Qianyu He\textsuperscript{1}, Jiaqing Liang\textsuperscript{2\textdagger} \\
\textbf{Weikang Zhou\textsuperscript{3}, Zeye Sun\textsuperscript{3}, Fei Yu\textsuperscript{3}, Yanghua Xiao\textsuperscript{1\textdagger}}
    \\
    \textsuperscript{1}Shanghai Key Laboratory of Data Science, School of Computer Science, Fudan University \\
    \textsuperscript{2}School of Data Science, Fudan University  \textsuperscript{3}Ant Group\\
    \{qxhe23, qyhe21\}@m.fudan.edu.cn, \{liangjiaqing, shawyh\}@fudan.edu.cn\\
}

\begin{document}
\maketitle
\begin{abstract}
Logical reasoning is essential for large language models (LLMs) to ensure accurate and coherent inference. However, LLMs struggle with reasoning order variations and fail to generalize across logically equivalent transformations. LLMs often rely on fixed sequential patterns rather than true logical understanding. To address this issue, we introduce an order-centric data augmentation framework based on commutativity in logical reasoning. We first randomly shuffle independent premises to introduce condition order augmentation. For reasoning steps, we construct a directed acyclic graph (DAG) to model dependencies between steps, which allows us to identify valid reorderings of steps while preserving logical correctness. By leveraging order-centric augmentations, models can develop a more flexible and generalized reasoning process. Finally, we conduct extensive experiments across multiple logical reasoning benchmarks, demonstrating that our method significantly enhances LLMs' reasoning performance and adaptability to diverse logical structures. We release our codes and augmented data in \url{https://github.com/qianxiHe147/Order-Centric-Data-Augmentation}.
\end{abstract}

\renewcommand{\thefootnote}{\textdagger}
\footnotetext{Corresponding authors.}

\section{Introduction}

Large language models (LLMs) have demonstrated exceptional performance across various real-world applications~\cite{jaech2024openai,dubey2024llama,liu2024deepseek}. Logic reasoning~\cite{Cummins1991-CUMCRA-2} is essential for LLMs. It allows models to draw valid conclusions, maintain coherence, and make reliable decisions across tasks~\cite{pan2023logic,liu2023evaluating}.

However, LLMs are sensitive to reasoning order and struggle with logically equivalent transformations~\cite{chen2024premise,berglund2023reversal,tarski1956logic}. First, the models are highly sensitive to the order of premises, with perturbing the order leading to up to a 40\% performance drop~\cite{chen2024premise,liu2024conciseorganizedperceptionfacilitates}. Additionally, if the testing order is reversed compared to the training order, accuracy drops drastically. For example, in the case of data involving two entities within a single factual statement, accuracy drops from 96.7\% to 0.1\% when training is left-to-right and testing is right-to-left.~\cite{berglund2023reversal,berglund2023taken,allen2023physics}. This suggests that LLMs follow a rigid logical reasoning order driven by learned patterns rather than true logical understanding.

\begin{figure}[t] 
    \centering
        \includegraphics[width=0.45\textwidth]{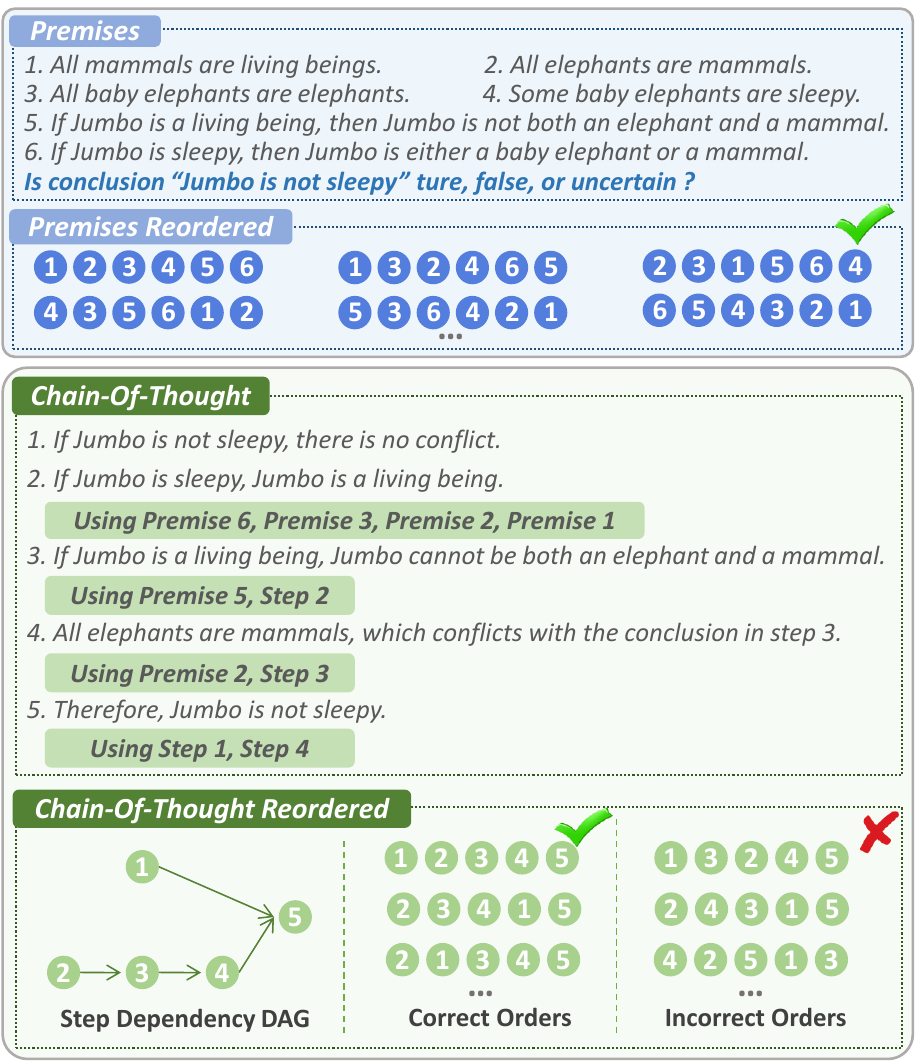}
    \caption{A logical reasoning example. Independent premises can be freely reordered, while reasoning steps must be reordered without violating dependencies.}
    \label{fig:order_intro}
\end{figure}

Existing LLM logical data augmentation methods do not effectively address the sensitivity to equivalent transformations. First, many logical datasets are specifically designed for certain domains, such as specialized fields or exam questions, primarily to broaden the scope of logical reasoning data collection and application~\cite{han2022folio,liu2020logiqa,yu2020reclor}. Second, a line of work aims to enhance the model's reasoning by mapping natural language to symbolic reasoning~\cite{olausson2023linc,xu2024faithful,pan2023logic}, but it primarily provides symbolic tools for understanding logical language rather than enhancing the logical structure itself. Lastly, another augmentation method creates a ``vacuum'' world to block interference from real-world logic~\cite{saparov2022language}, but it focuses on the impact of the model’s prior experience on reasoning, without addressing the design of logical equivalence.

In fact, \textbf{commutativity} is a fundamental property of logical reasoning, ensuring that independent logical units can be reordered without altering their meaning.  As established by Gödel’s completeness theorem~\cite{godel1930completeness} and Tarski’s model theory~\cite{tarski1956logic}, commutativity means that independent logical units can be freely reordered without changing the essence of the logical structure. Therefore, in logical reasoning, first, independent premises are commutative. As shown in the upper half of Fig. \ref{fig:order_intro}, different orders of premises represent equivalent problem structures. Furthermore, as demonstrated by Gentzen’s proof theory~\cite{gentzen1935proof}, reasoning steps are also commutative, provided their dependencies are intact. As shown in the lower half of Fig. \ref{fig:order_intro}, changing the order of steps without disrupting the dependencies results in an equivalent reasoning process. However, altering the order of dependent steps disrupts inference and prevents a coherent path to the correct conclusion.

In this work, we propose an order-centric data augmentation framework that explicitly incorporates logical commutativity into LLM training. For condition order, we randomly shuffle all independent premises. For reasoning steps, we construct a structured, step-by-step reasoning process, identify step dependencies using a directed acyclic graph (DAG), and apply topological sorting to reorder reasoning steps while preserving logical dependencies. Order-centric data augmentation allows models to learn logical equivalence through commutativity, leading to a deeper understanding of logic, rather than relying solely on fixed patterns to solve problems. Our experiments show that order-centric augmentation outperforms training on datasets with a fixed logical structure, enhancing the model's overall reasoning ability and improving its performance in complex shuffled testing scenarios.

Our contributions are summarized as follows:
(1) We propose an order-centric logic data augmentation method based on commutativity, which permutes both condition order and reasoning step order, helping models gain a deeper understanding of logical equivalence.
(2) We introduce a method that uses DAGs to model the dependencies between reasoning steps, helping to identify valid step reorderings.
(3) We conduct extensive experiments to prove the effectiveness of our approach in enhancing logical reasoning.

\section{Related Work}
\begin{figure*}[t] 
    \centering
        \includegraphics[width=1\textwidth]{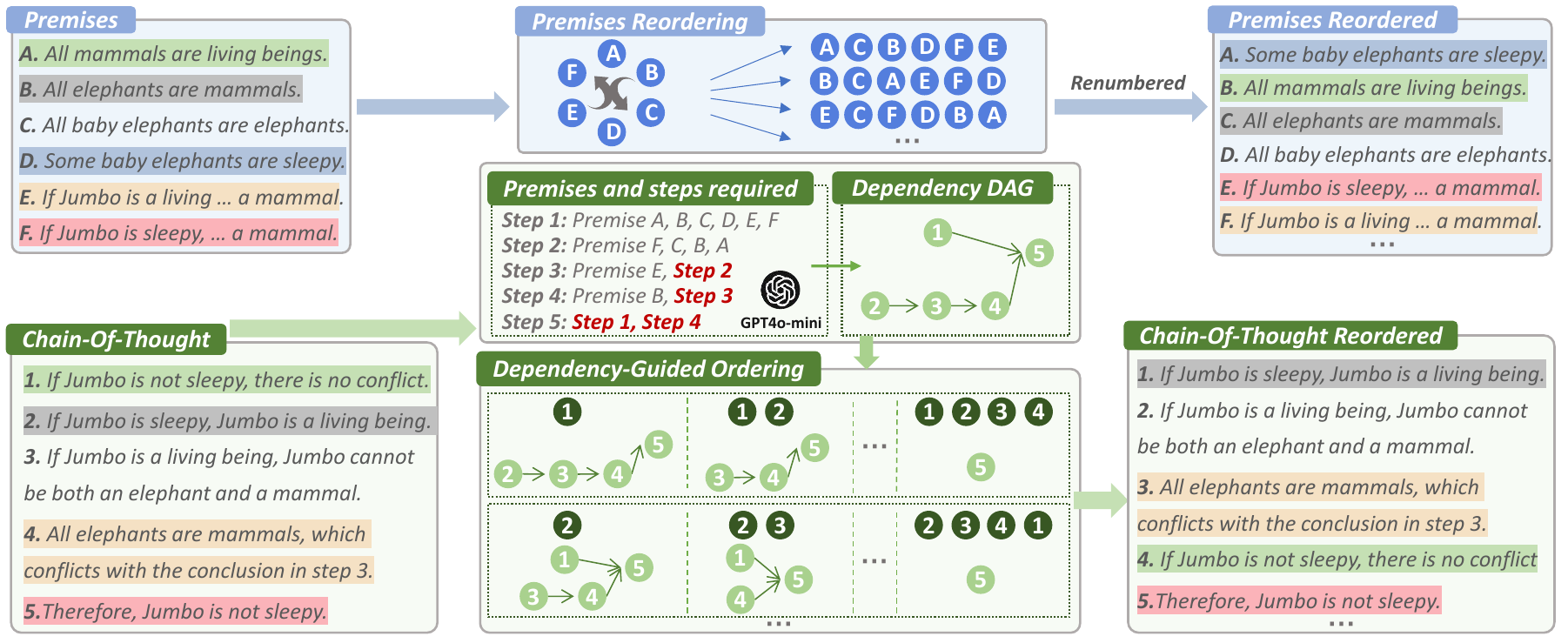}
    \caption{The framework of order-centric data augmentation method. First, we apply condition augmentation by randomly reordering independent premises. Then, we enhance reasoning step order through a directed acyclic graph (DAG) to identify step dependencies and reorder them while preserving logical correctness.}
    \label{fig:method}
\end{figure*}

\subsection{Order Effect of Language Models}
Large language models are sensitive to reasoning order (an example is shown in Appendix ~\ref{sec:appendix_premise_matter}). While word order variations in natural language have little impact~\cite{cao2023unnatural,abdou2022word}, disrupting the order in reasoning tasks significantly degrades performance. \citet{chen2024premise} show that models perform optimally only when the premise order matches the sequence required for the reasoning process. To address this, \citet{liu2024conciseorganizedperceptionfacilitates} propose reorganizing premise order to reduce order sensitivity. However, this approach is task-specific and lacks generalizability.
Furthermore, the Reversal Curse reveals that models fail to grasp logical equivalence when trained with a fixed linguistic order~\cite{berglund2023reversal}. \citet{golovneva2024reverse} mitigate this by proposing reverse training, where LLMs learn both forward and reverse reasoning by randomly shuffling words or segments within a sentence. This highlights the need for diverse training data with varied orderings.


Compared to the above works, we extend to more complex logical reasoning scenarios, building upon this concept by leveraging commutativity for data augmentation in logical reasoning, which helps models generalize across different reasoning structures and enhances robustness.

\subsection{Logical Reasoning Enhancing}
Existing methods to enhance LLMs' logical reasoning ability mainly fall into three categories: integrating symbolic reasoning, training and inference strategies, and leveraging data augmentation.

Symbolic reasoning enhances LLMs by transforming natural language into formal logic, providing a symbolic approach that helps models understand logic~\cite{olausson2023linc,xu2024faithful,zhang2023improved}. Training and inference strategies use adversarial pre-training, contrastive learning, and multi-step explicit planning to improve training efficiency and reasoning effectiveness~\cite{pi2022logiganlearninglogicalreasoning,jiao2022merit,zhao2023explicitplanninghelpslanguage}. 
Data augmentation creates diverse training and testing data, aiding models in generalizing better across logical tasks~\cite{han2022folio,tafjord2021proofwritergeneratingimplicationsproofs,clark2020transformerssoftreasonerslanguage}. 
LogiGLUE~\cite{luo2023towards} builds a large-scale logical benchmark through instruction fine-tuning across deductive, abductive, and inductive tasks.
LogicBench~\cite{parmar2024logicbench} focuses on single-rule inference, evaluating LLMs on 25 reasoning patterns and exposing their weaknesses in complex reasoning and negation handling.

Our work falls into the last category. Unlike the previous approaches, we perform order-centric data augmentation on existing logical reasoning datasets, leveraging logical commutativity to enhance the model's understanding of logical equivalence and improve overall logical reasoning ability.

\section{Problem Formulation}
In this paper, we formulate the problem of logical reasoning in a unified representation. Let \( D = \{P, C, L\} \) represent a logical reasoning problem, where \( P = \{P_1, P_2, \dots, P_n\} \) is the set of premises, \( C \) is the conclusion, and \( L \) is the label, which takes a value from a finite set, such as \( \{ \text{true}, \text{false}, \text{uncertain} \} \), indicating whether \( C \) can be logically inferred from \( P \). In step-based data augmentation, we extend the representation to include a solution \( S = \{S_1, S_2, \dots, S_m\} \), where \( S \) consists of reasoning steps that derive the conclusion from the premises. This process can be abstracted as a directed acyclic graph (DAG). Typical logical reasoning datasets only provide labels. Therefore, we construct \( S \) ourselves. The specific construction of \( S \) will be detailed in Sec. \ref{sec:Answer order Augmentation}.

\section{Method}


In this section, we introduce condition order augmentation in Sec. \ref{sec:Condition Order Augmentation} and answer order augmentation in Sec. \ref{sec:Answer order Augmentation}. The framework is shown in Fig. \ref{fig:method}.

\subsection{Condition Order Augmentation}
\label{sec:Condition Order Augmentation}
Due to the commutativity of premises, swapping independent premises results in the same solution. Hence, we perturb the order of premises, enabling models to learn the logical equivalence of condition reordering.

\subsubsection{Shuffling the Order of Premises}
Given a logical reasoning dataset \( D_C = \{P, C, L\} \), we first extract the premise set \( P = \{P_1, P_2, \dots, P_n\} \). To generate augmented data, we apply a random permutation \( \sigma \) to the premise set \( P \), producing a new ordered premise set \( P_{ran} \). Specifically:
\[ P_{ran} = \{P_{\sigma(1)}, P_{\sigma(2)}, \dots, P_{\sigma(n)}\} \]
For example, if the original order is \( [P_1, P_2, P_3, P_4, \dots, P_n] \), after applying the permutation \( \sigma \), the new order might be \( [P_3, P_4, P_1, P_n, \dots, P_2, \dots] \).

\subsubsection{Generating Augmented Data}
We denote the original dataset as \( D_C = \{P, C, L\} \) and the augmented dataset as \( D_C' = \{P_{ran}, C, L\} \), where \( P_{ran} \) represents the randomly shuffled premises. The transformation from \( D_C \) to \( D_C' \) involves perturbing the order of the premises while keeping the conclusion \( C \) and the label \( L \) unchanged. Each original data sample generates \( k \) instances of condition order augmentation, leading to an augmented dataset \( D_C' \) containing \( k \times |D_C| \) instances, where \( |D_C| \) is the size of the original dataset.


\subsection{Answer order Augmentation}
\label{sec:Answer order Augmentation}
Due to the commutativity of reasoning steps, we perturb the order of solution steps to help models learn the logical equivalence of the reasoning process. However, reasoning steps often have dependencies, where the execution of one step may rely on the result of another. To address this, we propose a method for identifying valid step reorderings that ensures these dependencies are preserved.

\begin{figure}[t] 
    \centering
        \resizebox{0.48\textwidth}{!}{
            \includegraphics[width=1\textwidth]{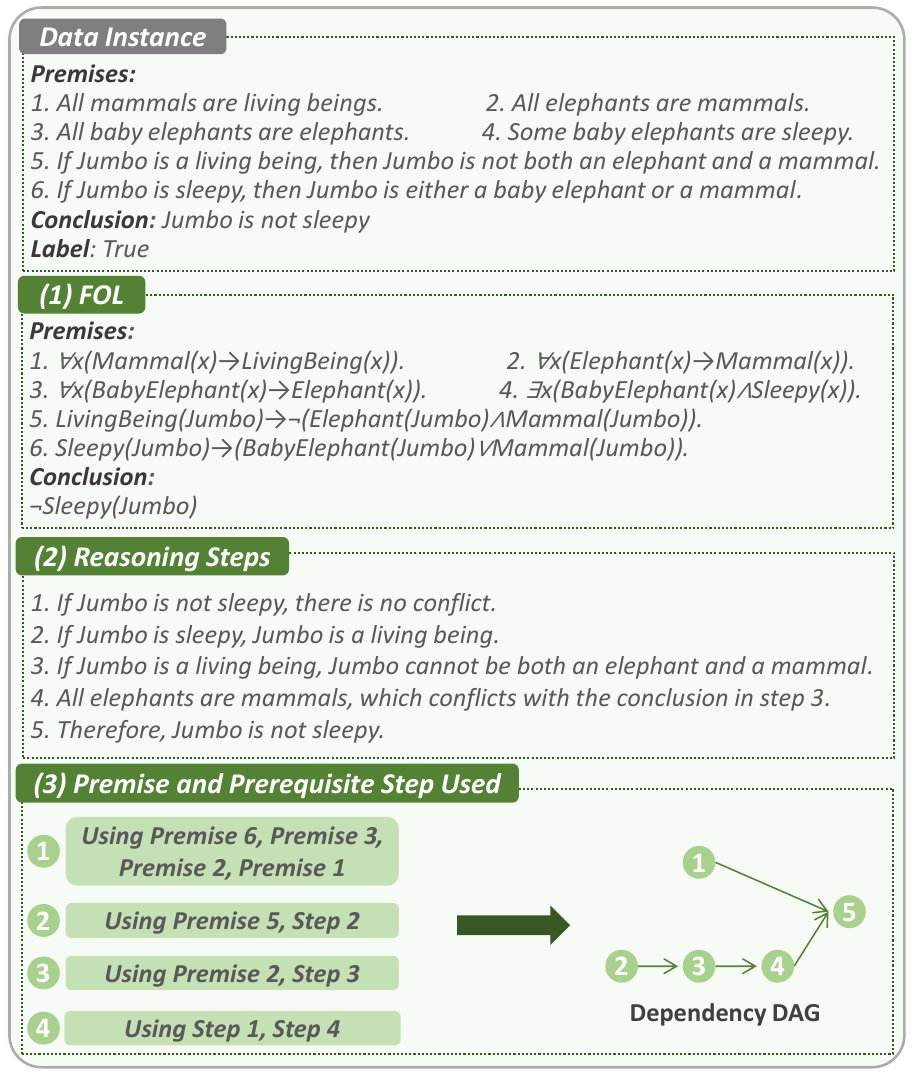}
    }
    \caption{An example of generating a specific solution from data containing only labels and constructing a Directed Acyclic Graph (DAG) to represent the dependencies between steps. Due to space limitations, we only list the conclusions of each step without showing the detailed content.}
    \label{fig:solution_example}
\end{figure}
\subsubsection{Leveraging LLMs for Logical Reasoning Solutions}
Since logical reasoning datasets typically provide only a single label (e.g., true/false) without a Chain-of-Thought (CoT) reasoning process, we generate detailed step-by-step reasoning solutions to bridge this gap~\cite{xu2024faithful}. We use LLMs$\footnote{In our experiment, we use GPT-4o-mini.}$ for this process. As shown in Fig. \ref{fig:solution_example}, the methodology consists of three main steps:
(1) For datasets without First-Order Logic (FOL) expressions, We extract their premises and conclusion and convert them into the corresponding FOL representations.
(2) The FOL-augmented premises, along with the ground truth labels, are fed into the model, prompting it to generate a step-by-step solution. Each step must clarify its purpose and reasoning, leading to a final conclusion.
(3) The generated solutions are then reprocessed by the model to extract the premise indices and prerequisite step indices used in each reasoning step.

\subsubsection{Constructing the Step Dependency DAG}
After obtaining the logical reasoning solutions, the current data can be represented as \( D_S = \{P, C, L, S\} \), where \( S = \{S_1, S_2, \dots, S_m\} \) consists of reasoning steps.
We represent \( S \) as a directed acyclic graph (DAG), denoted as \( G = (V, E) \), where \( V = \{S_1, S_2, \dots, S_m\} \) is the set of reasoning steps, and \( E \subseteq V \times V \) is the set of directed edges. An edge \( (S_i, S_j) \) indicates that step \( S_j \) depends on the result of step \( S_i \).

Each step \( S_i \) is represented as a tuple:
\[
S_i = (\text{Goal}_i, \mathcal{P}_{used}^{(i)}, \mathcal{S}_{used}^{(i)}, \text{Result}_i)
\]
where \( \text{Goal}_i \) describes the goal of the step, \( \mathcal{P}_{used}^{(i)} \) represents the directly used atomic premises, \( \mathcal{S}_{used}^{(i)} \subseteq V \) denotes the prerequisite steps that must be executed before \( S_i \), and \( \text{Result}_i \) is the result derived from the execution of \( S_i \). 

\subsubsection{Generating Augmented Solution Sequences}
A valid reasoning process must maintain all logical dependencies between steps while allowing flexibility in ordering interchangeable steps. We define the dependency constraints as follows:
\begin{itemize}
    \item A step \( S_i \) is \textbf{independent} if \( \mathcal{S}_{used}^{(i)} = \emptyset \) (i.e., it has no prerequisite steps).
    \item A step \( S_j \) is \textbf{dependent} if \( \mathcal{S}_{used}^{(i)} \neq \emptyset \), meaning that it requires prior steps to be completed before execution.
    \item Two steps \( S_i \) and \( S_j \) are \textbf{order-invariant} if neither step appears in the other's prerequisite set, i.e., \( S_i \notin \mathcal{S}_{used}^{(j)} \) and \( S_j \notin \mathcal{S}_{used}^{(i)} \).
\end{itemize}

Our goal is to generate a reasoning sequence that integrates all steps while maintaining dependency constraints, based on the principles outlined above. We represent the dependency graph as a dictionary, where each key is a step name and its value is a list of steps it depends on. First, we identify all independent steps where \( \mathcal{S}_{used}^{(i)} = \emptyset \) from the dataset, remove them from the DAG, and add them to the list of feasible step sequences \textit{List}. Then, we iterate over all possible combinations of these steps to generate multiple different lists of valid sequences. Next, for each step still present in the DAG, we iterate through the steps of its \( \mathcal{S}_{used} \). If \( \mathcal{S}_{used} \) contains a step from the current \textit{List}, we remove that step from \( \mathcal{S}_{used} \).

We repeat these two steps until every \textit{\(\text{List}\)} contains all the steps from \( S \), resulting in a collection of new valid step sequences. We refer to each newly generated sequence as \( S_{ran} \). The final augmented dataset is represented as \( D_S'
 = \{P, C, L, S_{ran}\} \).

Details of the prompts used in order-centric data augmentation are provided in Appendix \ref{sec:appendix_1}.

\section{Experiments}
We conduct experiments to evaluate the effectiveness of our method, focusing on overall performance, training efficiency, and generalization capability.

\subsection{Experiment Setup}
\paragraph*{Datasets} 
(1) \textbf{FOLIO}~\cite{han2022folio} is a natural language inference dataset annotated with first-order logic (FOL), consisting of 1001 training samples and 231 test samples.  
(2) \textbf{RuleTaker}~\cite{clark2020transformerssoftreasonerslanguage} requires models to determine whether a conclusion is entailed by a set of premises, covering various reasoning difficulties. Due to its large scale, we uniformly sample 1000 training and 1000 test instances across different difficulty levels.  
(3) \textbf{LogicNLI}~\cite{tian-etal-2021-diagnosing} is an NLI-style dataset that isolates first-order logic reasoning from commonsense inference for precise logical evaluation. Similarly, we sample 1000 instances from both its training and test sets.  

\paragraph*{Models} 
We conduct experiments on Llama-3-8B-Instruct~\cite{llama3modelcard}, Llama-2-13B-Chat~\cite{touvron2023llama} and Mistral-7B-Instruct-v0.3~\cite{jiang2023mistral}, evaluating model performance under five training conditions:  
(1) \textbf{Untrained}: The original model without any additional training.  
(2) \textbf{Vanilla SFT}: Models fine-tuned only on the original training set, i.e., \( D_C = \{P, C, L\} \). 
(3) \textbf{Vanilla SFT + Condition Shuffling}: Models trained on both the original dataset and an augmented version with shuffled condition orders, i.e., \( D_C = \{P, C, L\} \) and \( D_C' = \{P_{ran}, C, L\} \). 
(4) \textbf{SFT with COT}: Models fine-tuned with training data that includes Chain-of-Thought (COT) reasoning steps, i.e., \( D_S
 = \{P, C, L, S\} \).  
(5) \textbf{SFT with COT + Answer Steps Shuffling}: A model trained with COT data and additional augmentations with shuffled reasoning steps, i.e., \( D_S
 = \{P, C, L, S\} \) and \( D_S'
 = \{P, C, L, S_{ran}\} \).

All models are trained using full fine-tuning, with a 1:1 mix of ShareGPT~\cite{chiang2023vicuna} in each dataset. Training is conducted on four A100 GPUs for four epochs. Each model is trained exclusively on a single dataset, with augmentation applied only to that dataset, and evaluated on the corresponding test set without cross-dataset mixing.


\begin{table}[!htb]
\renewcommand{\arraystretch}{1.2} 
\renewcommand{\familydefault}{\rmdefault}
    \centering
    \resizebox{0.48\textwidth}{!}{
\begin{tabular}{llccc}
\toprule
\textbf{Set} & \textbf{Type} & \textbf{FOLIO} & \textbf{RuleTaker} & \textbf{LogicNLI} \\
\hline
\multirow{3}{*}{Train} & Original             & 1001  & 1000      & 1000     \\
& Condition Shuffled   & 1001  & 1000      & 1000     \\
& Answer Step Shuffled & 619   & 594       & 627      \\
\hdashline
\multirow{2}{*}{Test}  & Original             & 203   & 2000      & 2000     \\
& Condition Shuffled   & 406   & 2000      & 2000   \\
\bottomrule 
\end{tabular}
    }
    \caption{The data sizes of the training and test sets used in the main experiments.}
    \label{tab:Data_statistics}
\end{table}
We applied random shuffling to the premises of each training sample to generate one condition-augmented instance. Due to some data containing multiple valid step orderings, we randomly selected one transformation from each original data to control the data size. Additionally, we shuffled the premises in the test set to create a condition shuffled test set, enabling better evaluation of the model's performance across different logical orders. The data sizes for both the training and test sets are provided in Tab. \ref{tab:Data_statistics}.

\begin{table*}[t]
\newcolumntype{g}{>{\columncolor{green!4}}c}
\newcolumntype{b}{>{\columncolor{blue!4}}c}
\renewcommand{\arraystretch}{1.2} 
\renewcommand{\familydefault}{\rmdefault}
\resizebox{\textwidth}{!}{
\begin{tabular}{llccccccgb}
\toprule
\multirow{2}{*}{\textbf{Models}}        & \multirow{2}{*}{\textbf{Training}} & \multicolumn{2}{c}{\textbf{FOLIO}}    & \multicolumn{2}{c}{\textbf{RuleTaker}}  & \multicolumn{2}{c}{\textbf{LogicNLI}}   & \multicolumn{2}{c}{\textbf{Avg.}} \\
\cmidrule(lr){3-10}
&             & \textbf{Seq.} & \textbf{Shf.} & \textbf{Seq.} & \textbf{Shf.} & \textbf{Seq.} & \textbf{Shf.} & \textbf{Seq.} & \textbf{Shf.} \\
\hline
\multirow{5}{*}{LLaMA3-8B-Instruct}     & Untrained   & 57.64\% & 55.17\% & 57.95\% & 57.95\% & 28.85\% & 25.50\% & 48.15\%   & 46.21\%           \\
& Vanilla SFT & 63.55\% & 61.82\% & 70.65\% & 68.65\% & 54.40\%  & 54.90\% & 62.87\%   & 61.79\%            \\
& + Condition Shuffled  & 70.44\%\textcolor{darkgreen}{(+6.89)} & 67.49\%\textcolor{darkgreen}{(+5.67)} & 81.70\%\textcolor{darkgreen}{(+11.05)} & 80.15\%\textcolor{darkgreen}{(+11.50)} & 59.95\%\textcolor{darkgreen}{(+5.55)} & 59.45\%\textcolor{darkgreen}{(+4.55)} & 70.70\%\textcolor{darkgreen}{(+7.83)}   & 69.03\%\textcolor{darkgreen}{(+7.24)}           \\
& SFT with COT  & 76.35\% & 73.65\% & 81.05\% & 78.80\% & 42.20\% & 41.65\% & 66.53\%   & 64.70\%  \\
& + Answer Steps Shuffled & 77.34\%\textcolor{darkgreen}{(+0.99)} & 76.85\%\textcolor{darkgreen}{(+3.20)} & 84.60\%\textcolor{darkgreen}{(+3.55)} & 82.70\%\textcolor{darkgreen}{(+3.90)} & 43.80\%\textcolor{darkgreen}{(+1.60)} & 42.80\%\textcolor{darkgreen}{(+1.15)} & 68.58\%\textcolor{darkgreen}{(+2.05)}   & 67.45\%\textcolor{darkgreen}{(+2.75)}           \\
\hdashline
\multirow{5}{*}{LLaMA2-13B-Chat}        & Untrained   & 39.16\% & 35.47\% & 53.30\% & 52.50\% & 28.45\% & 26.95\% & 40.30\%   & 38.31\%           \\
& Vanilla SFT & 53.69\% & 51.71\% & 65.35\% & 62.80\% & 45.65\% & 44.20\% & 54.90\%   & 52.90\%           \\
& + Condition Shuffled  & 63.05\%\textcolor{darkgreen}{(+9.36)} & 62.32\%\textcolor{darkgreen}{(+10.61)} & 72.20\%\textcolor{darkgreen}{(+6.85)} & 71.30\%\textcolor{darkgreen}{(+8.50)} & 54.00\%\textcolor{darkgreen}{(+8.35)} & 54.10\%\textcolor{darkgreen}{(+9.90)} & 63.09\%\textcolor{darkgreen}{(+8.19)}   & 62.57\%\textcolor{darkgreen}{(+9.67)}           \\
& SFT with COT  & 73.40\% & 70.69\% & 76.60\% & 74.65\% & 43.70\% & 39.30\% & 64.57\%   & 61.55\%           \\
& + Answer Steps Shuffled & 76.35\%\textcolor{darkgreen}{(+2.95)} & 73.89\%\textcolor{darkgreen}{(+3.20)} & 75.50\%\textcolor{red}{(-1.10)} & 72.25\%\textcolor{red}{(-2.40)} & 46.90\%\textcolor{darkgreen}{(+3.20)} & 42.75\%\textcolor{darkgreen}{(+3.45)} & 66.25\%\textcolor{darkgreen}{(+1.68)}   & 62.97\%\textcolor{darkgreen}{(+1.42)}         \\
\cdashlinelr{1-10} 
\multirow{5}{*}{Mistral-7B-Instruct-v0.3} & Untrained   & 56.16\% & 57.39\% & 54.55\% & 54.55\% & 26.55\% & 24.90\% & 45.75\%   & 45.61\%           \\
& Vanilla SFT & 54.68\% & 55.17\% & 52.90\% & 53.20\% & 25.15\% & 25.05\% & 44.24\%   & 44.47\%           \\
& + Condition Shuffled  & 62.07\%\textcolor{darkgreen}{(+7.39)} & 61.82\%\textcolor{darkgreen}{(+6.65)} & 70.95\%\textcolor{darkgreen}{(+18.05)} & 69.20\%\textcolor{darkgreen}{(+16.00)} & 43.00\%\textcolor{darkgreen}{(+17.85)} & 44.20\%\textcolor{darkgreen}{(+19.15)} & 58.67\%\textcolor{darkgreen}{(+14.43)}   & 58.41\%\textcolor{darkgreen}{(+13.94)}          \\
& SFT with COT  & 67.47\% & 66.75\% & 81.55\% & 77.00\% & 46.40\% & 44.85\% & 65.14\%   & 62.87\%           \\
& + Answer Steps Shuffled & 72.91\%\textcolor{darkgreen}{(+5.44)} & 72.17\%\textcolor{darkgreen}{(+5.42)} & 84.10\%\textcolor{darkgreen}{(+2.55)} & 82.80\%\textcolor{darkgreen}{(+5.80)} & 47.35\%\textcolor{darkgreen}{(+0.95)} & 47.30\%\textcolor{darkgreen}{(+2.45)} & 68.12\%\textcolor{darkgreen}{(+2.98)}   & 67.42\%\textcolor{darkgreen}{(+4.55)}\\
\bottomrule
\end{tabular}
}
  \caption{
  The overall performance on FOLIO, RuleTaker, and LogicNLI, where the value in parentheses after each model's Condition Shuffled represents the improvement relative to Vanilla SFT, and the value in parentheses after each model's Answer Steps Shuffled represents the improvement relative to SFT with COT.
  }
  \label{tab:main_result}
\end{table*}
\subsection{Overall Performance}



Tab.~\ref{tab:main_result} shows that our method effectively improves model reasoning performance. Compared to Vanilla SFT, condition shuffling significantly enhances results across multiple datasets, with overall gains ranging from 7\% to 15\%. Notably, it improves not only robustness to perturbed input order but also accuracy on original sequential test sets, suggesting enhanced logical reasoning ability rather than mere order tolerance. Further incorporating answer step shuffling brings additional improvements over CoT training, contributing an average boost of 2\% to 3\%, highlighting the benefit of diversified reasoning paths.

While LLaMA models tend to perform slightly better on sequential evaluations (typically 2--3\% higher), our shuffling strategies help mitigate this gap.

\begin{figure}[t] 
    \centering
        \resizebox{0.4\textwidth}{!}{
            \includegraphics[width=1\textwidth]{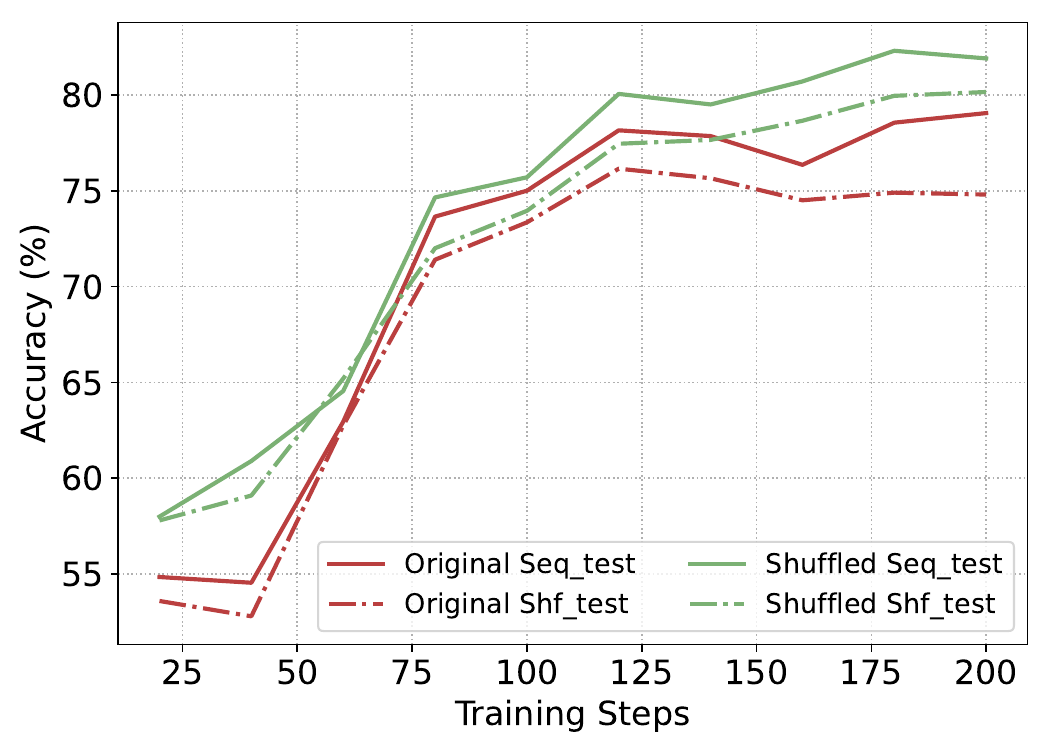}
    }
    \caption{The performance of training efficiency across different training steps in condition order augmentation.}
    \label{fig:condition_steps}
\end{figure}
\subsubsection{Training Efficiency}
To ensure fairness and exclude the effect of increased data size, we test the accuracy of checkpoints with the same number of training steps, comparing the performance of condition order augmentation with the original order. As shown in Fig. \ref{fig:condition_steps}, even with the same data size, condition shuffle training consistently outperforms the original order, with the performance gap widening as training progresses. This highlights that the improvement in accuracy is due to the augmentation process itself, rather than the increase in data.

\begin{table}[!htb]
\renewcommand{\arraystretch}{1.3} 
\renewcommand{\familydefault}{\rmdefault}
    \centering
    \resizebox{0.48\textwidth}{!}{
\begin{tabular}{llcc}
\toprule
\multirow{2}{*}{\textbf{Models}} & \multirow{2}{*}{\textbf{Training Data}} & \multicolumn{2}{c}{\textbf{Performance}} \\
\cmidrule(lr){3-4}
& & \textbf{GSM8K} & \textbf{MATH} \\
\hline
\multirow{3}{*}{LLaMA3-8B-Instruct} 
& mathematics\_trainset+FOLIO\_trainset & 65.28\% & 21.20\% \\
& + FOLIO Condition Shuffled & 65.81\% & 21.58\% \\
& + FOLIO Answer Steps Shuffled & \textbf{68.16\%} & \textbf{22.56\%} \\
\bottomrule
\end{tabular}
    }
    \caption{
    Performance comparison of LLaMA3-8B-Instruct on GSM8K and MATH benchmarks when trained with different augmentation strategies.
    }
    \label{tab:generalization}
\end{table}
\subsubsection{Generalization}
To validate whether our order-centric data augmentation on logical datasets can generalize to other reasoning domains, we conduct experiments on two mathematical datasets: GSM8K and MATH. Specifically, for these two datasets, we mix their training sets with the original FOLIO dataset, the condition-shuffled augmented dataset, and the answer-shuffled augmented dataset during training. The results in Tab.~\ref{tab:generalization}.

The results demonstrate that our method can effectively generalize to other reasoning tasks, improving their reasoning performance. We believe this provides promising evidence of the broader applicability of our approach.

We further analyze the impact of reasoning step counts and compare with prompt-based methods; see Appendix~\ref{sec:answer_steps} and~\ref{sec:prompt_based}.

\section{Analysis}
\begin{table*}[]
\renewcommand{\arraystretch}{1.2} 
\newcolumntype{b}{>{\columncolor{blue!4}}c}
\renewcommand{\familydefault}{\rmdefault}
\resizebox{\textwidth}{!}{
\begin{tabular}{llccccccccccb}
\toprule
\textbf{Model}                                     & \textbf{Test}       & \textbf{-1.0}    & \textbf{-0.8}    & \textbf{-0.6}         & \textbf{-0.4}    & \textbf{-0.2}    & \textbf{0.0}     & \textbf{0.2}           & \textbf{0.4}           & \textbf{0.6}     & \textbf{0.8}              & \textbf{Random}           \\
\hline
\multirow{2}{*}{LLaMA3-8B-Instruct}       & Sequential & 69.45\% & 80.00\% &  \underline{80.55\%} & 75.90\% & 64.25\% & 74.80\% & 69.65\%       & 73.15\%       & 74.40\% & 67.50\%          & \textbf{81.05\%} \\
& Shuffled   & 67.55\% & 77.90\% & \underline{77.95\%} & 74.85\% & 64.10\% & 72.90\% & 68.50\%       & 70.05\%       & 73.30\% & 66.20\%          & \textbf{78.80\%} \\
\hdashline
\multirow{2}{*}{LLaMA2-13B-Chat}          & Sequential  & 68.20\% & 65.25\% & 71.40\%       & 66.90\% & 70.60\% & 60.40\% & 71.65\%       & 70.35\%       & 72.50\% & \textbf{73.20\%} & \underline{72.20\%}    \\
& Shuffled  & 67.65\% & 63.65\% & 69.60\%       & 65.10\% & 69.40\% & 58.75\% & 69.00\%       & 68.75\%       & 69.55\% & \underline{70.60\%}    & \textbf{71.30\%} \\
\hdashline
\multirow{2}{*}{Mistral-7B-Instruct-v0.3} & Sequential  & 64.75\% & 64.80\% & 54.50\%       & 65.60\% & 69.05\% & 50.95\% & 68.95\%       & \underline{69.65\%} & 67.55\% & 54.65\%          & \textbf{70.95\%} \\
& Shuffled  & 64.80\% & 64.10\% & 55.55\%       & 66.25\% & 67.75\% & 51.15\% & \underline{67.90\%} & 66.70\%       & 67.40\% & 54.55\%          & \textbf{69.20\%}\\
\bottomrule
\end{tabular}
}
  \caption{
  The performance of the models on RuleTaker with conditionally shuffled premises at different tau values. Tau = 1 represents the original order, tau = -1 indicates a complete reversal, tau = 0 means uniform shuffling, and "Random" refers to a fully random shuffle.
  }
  \label{tab:tau}
\end{table*} 
\subsection{Condition Augmentation with Varying Shuffling Degrees}
To investigate the effects of premise order transformations, we divide the Kendall tau distance \( \tau \) between different premise orders and the original order into 10 groups, each spanning a 0.2 range within [-1,1). A \( \tau \) value of 1 indicates forward order, -1 indicates a complete reversal, and 0 represents  a more uniform shuffling. Additionally, random shuffle means that \( \tau \) values from the entire range may be included. We conduct experiments on RuleTaker using different \( \tau \) values for condition-based data augmentation.

As shown in Tab. \ref{tab:tau}, random shuffling provides the best performance across all \( \tau \) values. The level of perturbation in premise order significantly affects model accuracy,  with differences exceeding 10\%. LLaMA3-8B-Instruct excels with negative \( \tau \) values, while LLaMA2-13B-Chat and Mistral-7B-Instruct-v0.3 do better with positive \( \tau \) values. Random shuffled in training data achieves the best overall performance, emphasizing the value of diverse data augmentation for more flexible and robust models.

\begin{table*}[]
\renewcommand{\arraystretch}{1.3} 
\renewcommand{\familydefault}{\rmdefault}
\resizebox{\textwidth}{!}{
\begin{tabular}{llcccccc}
\toprule
\multirow{2}{*}{\textbf{Models}}            & \multirow{2}{*}{\textbf{Training data}} & \multicolumn{2}{c}{\textbf{FOLIO}} & \multicolumn{2}{c}{\textbf{Ruletaker}} & \multicolumn{2}{c}{\textbf{LogicNLI}} \\
\cmidrule(lr){3-8}
&                         & \textbf{Sequential}   & \textbf{Shuffled}   & \textbf{Sequential}     & \textbf{Shuffled}     & \textbf{Sequential}     & \textbf{Shuffled}    \\
\hline
\multirow{2}{*}{LLaMA3-8B-Instruct}
& Answer Steps Shuffled    & 77.34\%      & 76.85\%    & 84.60\% & 82.70\%      & 43.80\% & 42.80\%     \\
& Random Steps Shuffled & 76.85\%(-0.49) & 69.21\%(-7.64) & 82.10\%(-2.50) & 81.20\%(-1.50) & 45.20\%\textcolor{darkgreen}{(+1.40)} & 42.75\%(-0.05) \\
& Condition\&Answer Shuffled 
& 74.88\% (-2.46) 
& 75.86\% (-0.99) 
& 81.15\% (-3.45) 
& 79.60\% (-3.10) 
& 42.80\% (-1.00) 
& 43.50\% (+0.70) \\
\hdashline
\multirow{2}{*}{LLaMA2-13B-Chat} 
& Answer Steps Shuffled    & 76.35\%      & 73.89\%    & 75.50\% & 72.25\%      & 46.90\% & 42.75\%     \\  
& Random Steps Shuffled & 71.92\%(-4.43) & 69.21\%(-4.68) & 74.75\%(-0.75) & 72.75\%\textcolor{darkgreen}{(+0.50)} & 43.70\%(-3.20) & 42.45\%(-0.30) \\
& Condition\&Answer Shuffled 
& 70.94\% (-5.41) 
& 67.24\% (-6.65) 
& 77.40\% (+1.90) 
& 73.80\% (+1.55) 
& 44.20\% (-2.70) 
& 41.60\% (-1.15) \\
\hdashline
\multirow{2}{*}{Mistral-7B-Instruct-v0.3} 
& Answer Steps Shuffled    & 72.91\%      & 72.17\%    & 84.10\% & 82.80\%      & 47.35\% & 47.30\%     \\
& Random Steps Shuffled & 71.43\%(-1.48) & 72.41\%\textcolor{darkgreen}{(+0.24)} & 82.95\%(-1.15) & 79.95\%(-2.85) & 44.75\%(-2.60) & 45.25\%(-2.05) \\
& Condition\&Answer Shuffled 
& 70.94\% (-1.97) 
& 70.44\% (-1.73) 
& 82.55\% (-1.55) 
& 81.70\% (-1.10) 
& 41.50\% (-5.85)
& 42.00\% (-5.30) \\
\midrule
\multicolumn{8}{c}{\textit{Condition\&Answer
Shuffled on Stronger LLM}} \\
\midrule
\multirow{2}{*}{Llama-3.3-70B-Instruct} 
& Answer Steps Shuffled    & \textbf{80.30\%}      & 78.08\%    & 79.90\% & 80.00\%      & 51.05\% & 49.65\%     \\
& Condition\&Answer Shuffled 
& 78.82\% (-1.48) 
& \textbf{79.80\%} \textcolor{darkgreen}{(+1.72)}
& \textbf{83.95\%} \textcolor{darkgreen}{(+4.05)}
& \textbf{82.65\%} \textcolor{darkgreen}{(+2.65)}
& \textbf{52.00\%} \textcolor{darkgreen}{(+0.95)}
& \textbf{51.40\%} \textcolor{darkgreen}{(+1.75)}\\
\bottomrule
\end{tabular}
}
  \caption{
  The performance of three different augmentation methods: the first row represents the original DAG-based Answer Steps Shuffled augmentation, the second row represents random step shuffling without dependencies in Sec. \ref{sec:non_DAG}, and the third row represents the combined condition and answer augmentation method in Sec. \ref{sec:both_ran}. The last two rows show our new results using Llama-3.3-70B-Instruct model.
  }
  \label{tab:DAG_or_not}
\end{table*}
\subsection{The Importance of DAG-based Step Dependency}
\label{sec:non_DAG}
To explore the importance of using DAG for step dependencies in step augmentation, we use the Answer Step Shuffled data from Tab. \ref{tab:Data_statistics} as a baseline. We randomly shuffle the steps in the original COT process and assess its performance to evaluate the impact of random step reordering without DAG dependencies.

As shown in Tab. \ref{tab:DAG_or_not}, not utilizing DAG dependencies leads to a performance drop compared to DAG-based augmentation. The decline is particularly severe on FOLIO, where LLaMA3-8B-Instruct and LLaMA2-13B-Chat show a drop of 7.64\% and 4.68\% in the shuffled test. In contrast, Ruletaker and LogicNLI experience a smaller decline. 

To explore the underlying cause of this phenomenon, we investigate the degree of dependency between steps in the step dependency DAG. We introduce the \textbf{Topological Freedom Index (TFI)}. This metric measures how loosely or tightly connected a DAG is, and it is calculated as follows:
\begin{equation}
TFI = \frac{\text{Number of valid sequences}}{\text{Factorial of the number of steps}}
\end{equation}

\begin{figure}[t] 
    \centering
        \resizebox{0.48\textwidth}{!}{
            \includegraphics[width=1\textwidth]{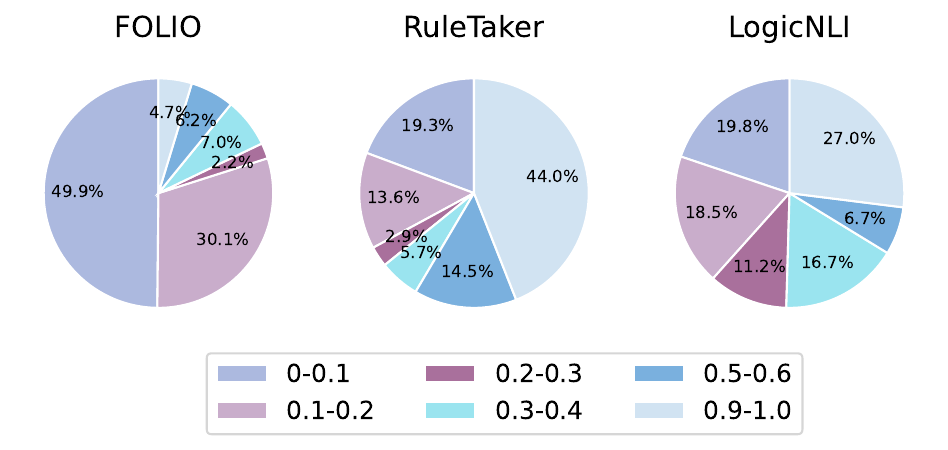}
    }
    \caption{The distribution of TFI index across different intervals in the training sets of FOLIO, RuleTaker, and LogicNLI. Since none of the datasets contain data in the [0.6-0.9) interval, this portion is omitted from the presentation.}
    \label{fig:pie_TFI}
\end{figure}

The number of valid sequences reflects the count of topological orderings consistent with the DAG’s dependencies, while the factorial of step count represents all possible orderings without constraints. A TFI value near 1 indicates weak dependencies and high reordering flexibility; a value near 0 suggests strong dependencies and strict sequencing. Fig.~\ref{fig:pie_TFI} shows the TFI distribution across three datasets, highlighting variations in step dependency among reasoning tasks.

In FOLIO, nearly half of the samples (49.9\%) have low TFI (0.0–0.1), indicating strong step dependencies. In contrast, RuleTaker and LogicNLI have more high-TFI samples (44.0\% and 27.0\% in the 0.9–1.0 range), reflecting looser reasoning structures. These differences align with our Random Step Shuffling results—datasets with stronger dependencies suffer more from shuffling. Thus, preserving dependency integrity is crucial when applying answer order augmentation.

\subsection{Combined Condition and Step Shuffling}
\label{sec:both_ran}


To investigate the combined effect of condition and step order perturbations, we apply premise shuffling on top of the Answer Steps Shuffled data, adjusting premise references in the answers accordingly. As shown in Table~\ref{tab:DAG_or_not}, \textbf{Condition\&Answer Shuffled} slightly underperforms compared to step-only shuffling in most cases.

This is likely because the two augmentations serve different purposes: condition shuffling teaches the model that independent premises are commutative, while step shuffling helps it understand that different reasoning paths under the same condition can lead to the same conclusion. When applied separately, each enhances logical equivalence learning; when combined, the dual perturbation increases learning difficulty, making it harder for smaller models to generalize.

To test this hypothesis, we evaluate the stronger LLaMA-3.3-70B-Instruct model. The results show a consistent 2–4\% improvement from joint perturbation over step-only shuffling on most datasets, suggesting that larger models are better equipped to handle complex logical variations and capture underlying equivalence.

\subsection{Effect of Augmentation Frequency}
In the main experiment, we set \( |D_C'| = |D_C| \), meaning that the parameter \( k = 1 \).
To investigate the impact of augmentation quantity, we increase \( k \) and generate \( k = 5 \) augmented instances for each original training sample in RuleTaker. 
This leads to an augmented dataset \( D_C' \) containing \( 5 \times |D_C| \) instances. 
As shown in Tab. \ref{tab:condition_variants}, adding a few shuffled instances improves model accuracy, but excessive augmentation results in performance degradation. This highlights the need to control the augmentation frequency. The increase in \( k \) can lead to a certain degree of performance improvement, indicating that our order-centric data augmentation method has room for further enhancement.

\begin{table}[t]
\renewcommand{\arraystretch}{1.2} 
\renewcommand{\familydefault}{\rmdefault}
    \centering
    \resizebox{0.48\textwidth}{!}{
\begin{tabular}{lccccc}
\toprule
Test       & \( k \)=1 & \( k \)=2 & \( k \)=3 & \( k \)=4 & \( k \)=5 \\
\hline
Sequential & 77.30\%    & 81.35\%     & \textbf{83.45\%}    & 83.15\%     & 77.65\%     \\
Shuffled   & 76.65\%    & 80.35\%     & 81.95\%    & \textbf{83.55\%}     & 79.85\%    \\
\bottomrule
\end{tabular}
    }
  \caption{
  The performance under different augmentation frequencies, where \( k \) represents the number of condition order augmentation instances applied per training sample.
  }
  \label{tab:condition_variants}
\end{table}

\section{Conclusion}

In this paper, we propose an order-centric data augmentation framework to improve LLMs' logical reasoning by addressing their sensitivity to reasoning order. By leveraging principles of independency and commutativity, we shuffle independent premises and use DAGs to guide valid step reorderings. Experiments across multiple benchmarks show that our method enhances reasoning performance and adaptability to diverse logical structures.

\section{Limitations}
Our work primarily focuses on logical commutativity within propositional reasoning tasks. However, this property extends beyond these tasks. It is also prevalent in many other reasoning scenarios, such as mathematical problems and other logic-based tasks. This remains an area for future exploration. Additionally, while we have explored the impact of condition order and answer order augmentations on model performance, how to further integrate and refine these augmentations for better logical reasoning capability is still an open question. We believe our exploration will provide valuable insights for future work on logical equivalence and commutativity in reasoning.

\bibliography{custom}
\clearpage

\appendix

\section{Appendix}
\label{sec:appendix}

\subsection{Premise Matter}
\label{sec:appendix_premise_matter}
From an overall performance perspective, in our main experiments, we observed that models generally perform worse on the shuffled testset compared to the sequential testset, as shown in the first row of results for each model in Tab.~\ref{tab:main_result}.

There is an examples in Tab.~\ref{tab:premise_matter} that illustrate the importance of premise order.
In the original order, the model correctly follows logical steps to reach the right conclusion. However, in the shuffled order, it fails to connect key premises, leading to an incorrect answer. This suggests that models may rely on surface-level cues rather than true logical understanding.
\begin{table*}[t]
\small
    \begin{tabularx}{\linewidth}{X}
    \toprule
\textbf{Original Premises:}

1. Beethoven leads the Vienna Music Society.

2. Beethoven wrote Symphony No. 9.

3. Composers write music pieces.

4. Vienna Music Society premiered Symphony No. 9.

5. Symphony No. 9 is a music piece.

6. Orchestras are led by conductors.

7. Vienna Music Society is an orchestra.

\textbf{Conclusion}: Some orchestras premiered music pieces.

\textbf{Label:} True

\textbf{Model's answer:}

\textbf{Step 1:} Identify the orchestra and the music piece. From premise 7, we know that Vienna Music Society is an orchestra: Orchestra(viennaMusicSociety) From premise 5, we know that Symphony No. 9 is a music piece: MusicPiece(symphony9)

\textbf{Step 2:} Premiere of the music piece by the orchestra. From premise 4, we know that Vienna Music Society premiered Symphony No. 9: Premiered(viennaMusicSociety, symphony9)

\textbf{Step 3:} Draw the conclusion. There exists an orchestra (Vienna Music Society) that premiered a music piece (Symphony No. 9). The conclusion is true.

\textbf{Shuffled Premises:}

1. Symphony No. 9 is a music piece.
2. Composers write music pieces.
3. Beethoven wrote Symphony No. 9.
4. Vienna Music Society premiered Symphony No. 9.
5. Vienna Music Society is an orchestra.
6. Beethoven leads the Vienna Music Society.
7. Orchestras are led by conductors.

\textbf{Model's answer:}

\textbf{Step 1:} Identify the orchestra and its leader. From premise 5, Vienna Music Society is an orchestra. From premise 6, Beethoven leads the Vienna Music Society.

\textbf{Step 2:} Analyze the relationship between orchestras and music pieces
Premise 3 states that composers write music pieces, but it does not mention orchestras premiering music pieces.

The conclusion claims "Some orchestras premiered music pieces," but the premises do not provide information about which orchestras premiered music pieces.

\textbf{Step 3:} Evaluate the conclusion
Since the premises do not provide sufficient information to support the claim that "Some orchestras premiered music pieces," the conclusion cannot be confirmed. \\

    \bottomrule
    \end{tabularx}
  \caption{
    An example where the order of premises affects the reasoning outcome.
  }
  \label{tab:premise_matter}
\end{table*}

\subsection{Details of Generating Solutions} 
\label{sec:appendix_1}
In Sec. \ref{sec:Answer order Augmentation}, 
We discuss how to generate step-by-step solutions through \( D = \{P, C, L\} \). Specifically, we follow these steps:

(1) For datasets that do not have first-order logic (FOL) expressions, such as RuleTaker and LogicNLI, we extract their premises and conclusions, and use GPT-4o-mini with prompts as shown in Tab. \ref{tab:100prompt_FOL} to convert them into corresponding FOL representations. FOLIO, on the other hand, already includes FOL expressions, so no conversion is required.

(2) The FOL-enhanced premises and ground truth labels are input into the model, prompting it to generate step-by-step solutions. As shown in the prompt in Tab. \ref{tab:100prompt_cot}, we add two domain-specific examples from each dataset to the prompt, requiring the model to clearly define the purpose and reasoning for each step, eventually leading to the final conclusion. The Task prompt specifies the possible values for the label. Specifically, in FOLIO, the label values are \{True, False, Unknown\}, in RuleTaker they are \{entailment, not entailment\}, and in LogicNLI they are \{entailment, neutral, self\_contradiction, contradiction\}.

(3) The model then reprocesses the generated solutions, using prompts like the one shown in Tab. \ref{tab:100prompt_DAG}, to extract the premise indices and premise step indices used in each reasoning step.

\subsection{Kendall Tau Distance}
In our study, we investigate the effects of premise order transformations by using the Kendall tau distance \( \tau \). This coefficient measures the correlation between two ordered lists, providing a quantitative way to assess how much one order differs from another. We use \( \tau \) to categorize various permutations of premise orders and assess their impact on model performance.

The Kendall tau coefficient \( \tau \) is calculated as follows:
\[
\tau = \frac{C - D}{\binom{n}{2}}
\]
where \( C \) is the number of concordant pairs (pairs of items that are in the same relative order in both lists), and \( D \) is the number of discordant pairs (pairs that are in opposite order in both lists). The total number of possible pairs is \( \binom{n}{2} \), where \( n \) is the number of items being compared.

We divide \( \tau \) values into 10 groups, each spanning a 0.2 range within the interval [-1, 1). A \( \tau \) value of 1 indicates that the order of the premises is exactly as required for the reasoning process, while -1 indicates a complete reversal of order. A \( \tau \) value of 0 indicates that the order is completely random, with no correlation to the original sequence.

\begin{figure}[t] 
    \centering
        \includegraphics[width=0.5\textwidth]{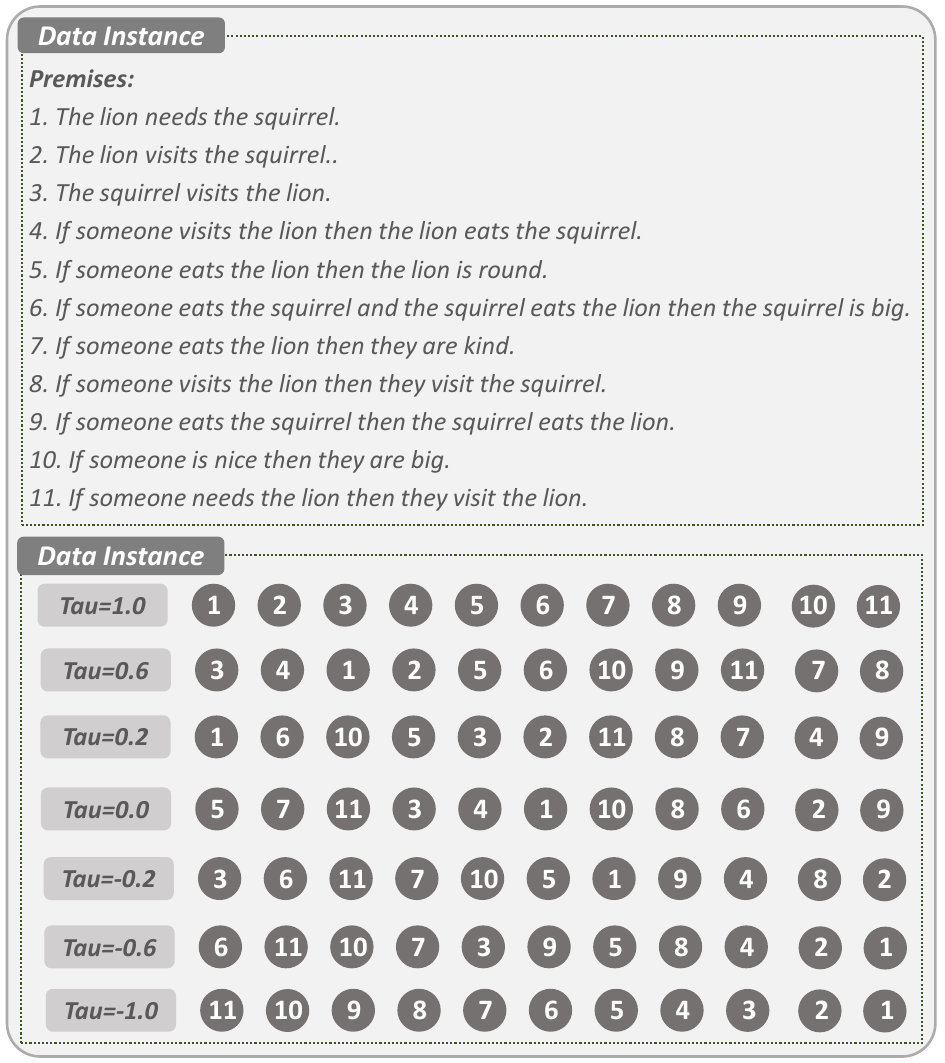}
    \caption{An example of showing the arrangement of premises with different tau values. The tau values do not represent exact values but rather the closest intervals for demonstration purposes.}
    \label{fig:tau}
\end{figure}

For example, if the original premise order is \( [P_1, P_2, P_3, \dots, P_n] \), a permutation function \( \sigma \) might rearrange it to \( [P_3, P_1, P_2, \dots, P_n] \). This process allows us to explore different levels of order perturbation, with the goal of analyzing how such variations affect model performance. Examples of premise orders corresponding to different \( \tau \) values can be seen in Fig. \ref{fig:tau}.


\begin{table}[!htb]
\normalsize
\renewcommand{\arraystretch}{1.5} 
\renewcommand{\familydefault}{\rmdefault}
    \centering
    \resizebox{0.48\textwidth}{!}{
\begin{tabular}{llcc}
\toprule
\textbf{Models} & \textbf{Training} & \textbf{LogicNLI Sequential} & \textbf{LogicNLI Shuffled} \\
\hline
\multirow{5}{*}{LLaMA3-8B-Instruct} & SFT with COT & 42.20\% & 41.65\% \\
& + Answer Steps Shuffled & 43.80\% & 42.80\% \\
& + 3\_Steps Shuffled & \textbf{51.9\%} & \textbf{50.15\%} \\
& + 4\_Steps Shuffled & 46.85\% & 44.95\% \\
& + 5\_Steps Shuffled & 41.3\% & 40.85\% \\
\hdashline
\multirow{5}{*}{Mistral-7B-Instruct-v0.3} & SFT with COT & 46.40\% & 44.85\% \\
& + Answer Steps Shuffled & 47.35\% & 47.30\% \\
& + 3\_Steps Shuffled & \textbf{52.25\%} & \textbf{51.10\%} \\
& + 4\_Steps Shuffled & 45.90\% & 41.30\% \\
& + 5\_Steps Shuffled & 41.85\% & 41.05\% \\
\bottomrule 
\end{tabular}
    }
    \caption{Performance on answer order augmentation LogicNLI subsets with different reasoning step lengths.}
    \label{tab:answer_steps}
\end{table}

\begin{table}[!htb]
\renewcommand{\arraystretch}{1.3} 
\renewcommand{\familydefault}{\rmdefault}
    \centering
    \resizebox{0.48\textwidth}{!}{
\begin{tabular}{llc}
\toprule
\textbf{Models} & \textbf{Method} & \textbf{FOLIO} \\
\hline
\multirow{2}{*}{GPT-4} 
& Logic-LM & 78.92\% \\
& SymbCoT & \textbf{82.33\%} \\
\hdashline
\multirow{3}{*}{GPT-3.5-turbo} 
& Logic-LM & \textbf{62.74\%} \\
& SymbCoT & 57.84\% \\
& COP & 47.00\% \\
\hdashline
\multirow{2}{*}{LLaMA3-8B-Instruct} 
& + Condition Shuffled & 70.44\% \\
& + Answer Steps Shuffled & \textbf{77.34\%} \\
\bottomrule
\end{tabular}
    }
    \caption{
    Performance comparison of different models and methods on the FOLIO benchmark. Our proposed Answer Steps Shuffled augmentation method achieves strong.
}
    \label{tab:folio_performance_comparison}
\end{table}
\subsection{Impact of Reasoning Step Length in Answer Order Augmentation}
\label{sec:answer_steps}
To investigate the impact of reasoning step count on our method, we conduct experiments using the LogicNLI dataset, which features a relatively balanced distribution of reasoning chains with different step lengths. Specifically, there are 647 instances suitable for answer shuffling, from which we select 150 samples with step lengths of 3, 4, and 5. These samples are combined with 1,000 original CoT samples to perform answer-shuffling-based data augmentation. The results on LLaMA3-8B-Instruct and Mistral-7B-Instruct-v0.3 are shown in Tab.~\ref{tab:answer_steps}.

The results indicate that using only Step-3 answer shuffling unexpectedly outperforms the original Answer Steps Shuffled method. Additionally, as the number of steps increases from 3 to 5, the model's performance gradually declines. This suggests that excessively complex reasoning step perturbations may negatively impact the model's ability to learn logical commutativity and equivalence.

\subsection{Comparison to Prompt\_Based Methods}
\label{sec:prompt_based}
To further validate the effectiveness of our approach, we conduct a comparative study against three representative symbolic reasoning methods: Logic-LM~\cite{pan2023logic}, SymbCoT~\cite{xu2024faithful}, and COP~\cite{liu2023concise}. These methods were primarily evaluated using proprietary large language models such as GPT-3.5-turbo and GPT-4. Among the datasets used in these works, FOLIO is included in our evaluation as well, enabling a direct performance comparison.

As shown in Tab. ~\ref{tab:folio_performance_comparison}, our method, despite being trained on the smaller open-source model LLaMA3-8B-Instruct, consistently outperforms GPT-3.5-turbo across all three symbolic reasoning baselines. Although it does not surpass GPT-4, our approach achieves competitive results given the substantial gap in model scale and computational resources. These findings underscore the efficiency and effectiveness of our training framework, particularly in enhancing logical reasoning capabilities without reliance on proprietary or excessively large models.

\begin{table*}[]
\small
    \begin{tabularx}{\linewidth}{X}
    \toprule
    \color{gray}{/* \textit{Task prompt} */}\\
    Please parse the context and question into First-Order Logic formulas. Please use symbols as much as possible to express, such as \( \forall \), \( \land \), \( \rightarrow \), \( \oplus \), \( \neg \), etc. 
    \\
    \color{gray}{/* \textit{Example} */}\\
\textbf{Premises:} \\
If a cartoon character is yellow, it is from the Simpsons. \\
If a cartoon character is from Simpsons, then it is loved by children. \\
Ben is ugly or yellow. \\
Ramon being real is equivalent to Rhett being not modest and Philip being lazy. \\

\textbf{Hypothesis:} \\
James does not have lunch in the company. \\

\textbf{Premises-FOL:} \\
\(\forall x (Yellow(x) \rightarrow Simpsons(x))\)

\(\forall x (Simpsons(x) \rightarrow Loved(x))\)

\((Yellow(ben) \vee Ugly(ben))\)

\(real(Ramon) \iff (modest(Rhett) \land lazy(Philip))\)

\textbf{Hypothesis-FOL:} \\
\(\neg HasLunch(james, company)\)
    \\
    \color{gray}{/* \textit{Input} */}\\
    ---INPUT---\\
    Premises:\\
    \{\textbf{Given\_premises}\}\\
    Hypothesis:\\
    \{\textbf{Given\_hypothesis}\}\\
    ---OUTPUT---\\
    \bottomrule
    \end{tabularx}
  \caption{
    The prompt for generating First-Order Logic (FOL) expressions corresponding to natural language logical propositions.
  }
  \label{tab:100prompt_FOL}
\end{table*}

\begin{table*}[t]
\small
    \begin{tabularx}{\linewidth}{X}
    \toprule
    \color{gray}{/* \textit{Task prompt} */}\\
    Please solve the question step by step based on First-Order Logic rules such as Modus Ponens, determine whether the hypothesis is true, false, or unknown based on these premises.
    \\
    \color{gray}{/* \textit{Example} */}\\
    \textbf{Premises:} \\
1. Walter Folger Brown was an American politician and lawyer who served as the postmaster general. \\
2. Walter Folger Brown graduated from Harvard University with a Bachelor of Arts. \\
3. While they were both in Toledo, Walter Folger Brown's father practiced law with Walter Folger Brown. \\
4. Katherin Hafer married Walter Folger Brown. \\

\textbf{Premises-FOL:} \\
1. AmericanPolitician(walterBrown) \(\land\) Lawyer(walterBrown) \(\land\) ServedAs(walterBrown, postMasterGeneral) \\
2. Graduated(walterBrown, harvard) \(\land\) GraduatedWith(walterBrown, bachelorsOfArt) \\
3. \(\exists t\) (In(walterBrown, toledo, t) \(\land\) In(walterBrownFather, toledo, t) \(\land\) PracticedLawTogether(walterBrown, walterBrownFather, t)) \\
4. Married(katherinHafer, walterBrown) \\

\textbf{Hypothesis:} \\
Walter Folger Brown was not in Toledo. \\

\textbf{Hypothesis-FOL:} \\
\(\exists t\) (\(\neg\)In(walterBrownFather, toledo, t)) \\
\textbf{Label:} \\
False \\
\textbf{Solution:} \\
\textbf{Step 1: Analyze Walter Folger Brown's presence in Toledo} \\
The third premise states that there exists a time t such that: \\
In(walterBrown, toledo, t) \(\land\) In(walterBrownFather, toledo, t) \(\land\) PracticedLawTogether(walterBrown, walterBrownFather, t). \\
This means that Walter Folger Brown and his father were both in Toledo at the same time, and they practiced law together there.\\
Thus, we have clear evidence that Walter Folger Brown was indeed in Toledo at some point. \\

\textbf{Step 2: Analyze the hypothesis's claim} \\
The hypothesis states that Walter Folger Brown was not in Toledo, represented in FOL as: \\
\(\exists t\) (\(\neg\)In(walterBrownFather, toledo, t)) \\
However, this contradicts the third premise, which explicitly states that both Walter Folger Brown and his father were in Toledo at the same time. \\
Therefore, the hypothesis that Walter Folger Brown was not in Toledo is False based on the premises. \\

\textbf{Final Hypothesis:} \\
The hypothesis "Walter Folger Brown was not in Toledo" is False.
    \\
    \color{gray}{/* \textit{Input} */}\\
    ---INPUT---\\
    Premises:\\
    \{\textbf{Given\_premises and premises-FOL}\}\\
    Hypothesis:\\
    \{\textbf{Given\_hypothesis and hypothesis-FOL}\}\\
    Label:\\
    \{\textbf{Given\_label}\}\\
    ---OUTPUT---\\
    \bottomrule
    \end{tabularx}
  \caption{
    The prompt for generating a step-by-step Chain of Thought (CoT) process based on premises, hypothesis, and label. Different datasets have different sets of labels and examples. For convenience, we only show the prompt on FOLIO here. In practice, we listed two examples, but for brevity and clarity in display, we only present one.
  }
  \label{tab:100prompt_cot}
\end{table*}

\begin{table*}[t]
\small
    \begin{tabularx}{\linewidth}{X}
    \toprule
    \color{gray}{/* \textit{Task prompt} */}\\
    I will provide you with a description of the question and its answer, and the condition of the question is specific. The answer is done in steps. I hope you can extract the conditions and prerequisite steps used in each step of the answer. Please note that I am not asking you to regenerate the answer yourself, but rather to extract the conditions and prerequisite steps used in each step from the answer I have given you. Meanwhile, the conditions used in the steps are quite explicit, but the prerequisite steps used are quite implicit. I hope you can understand and summarize the prerequisite steps used in each step. Your answer should only include Conditions and prerequisite steps used.
    \\
    \color{gray}{/* \textit{Example} */}\\
    \textbf{Question:} \\
\textbf{Premises:} \\
1. Lana Wilson directed After Tiller, The Departure, and Miss Americana. \\
2. If a film is directed by a person, the person is a filmmaker. \\
3. After Tiller is a documentary. \\
4. The documentary is a type of film. \\
5. Lana Wilson is from Kirkland. \\
6. Kirkland is a US city. \\
7. If a person is from a city in a country, the person is from the country. \\
8. After Tiller is nominated for the Independent Spirit Award for Best Documentary.

\textbf{Premises-FOL:} \\
1. DirectedBy(afterTiller, lanaWilson) \( \land \) DirectedBy(theDeparture, lanaWilson) \\
\( \land \) DirectedBy(missAmericana, lanaWilson) \\
2. \( \forall \) x \( \forall \) y (DirectedBy(x, y) \( \rightarrow \) Filmmaker(y)) \\
3. Documentary(afterTiller) \\
4. \( \forall \) x (Documentary(x) \( \rightarrow \) Film(x)) \\
5. From(lanaWilson, kirkland) \\
6. In(kirkland, unitedStates) \\
7. \( \forall \) x \( \forall \) y \( \forall \) z ((From(x, y) \( \land \) In(y, z)) \( \rightarrow \) From(x, z)) \\
8. Nomination(afterTiller, theIndependentSpiritAwardForBestDocumentary)

\textbf{Conclusion:} \\
Miss Americana is not directed by a filmmaker from Kirkland.

\textbf{Conclusion-FOL:}
\( \neg \) \(\exists t\) x (Filmmaker(x) \( \land \) From(x, kirkland) \( \land \) DirectedBy(missAmericana, x))
    \\

\textbf{Answer:} \\
Step 1: Analyze the premises regarding Lana Wilson's role as a filmmaker \\
From premise 1, we know that Lana Wilson directed Miss Americana. Therefore, we can conclude that Lana Wilson is a filmmaker based on premise 2, which states that if a film is directed by a person, that person is a filmmaker. Thus, we have:
    DirectedBy(missAmericana, lanaWilson) \( \rightarrow \) Filmmaker(lanaWilson)

Step 2: Analyze Lana Wilson's origin \\
From premise 5, we know that Lana Wilson is from Kirkland. Therefore, we can conclude:
    From(lanaWilson, kirkland)

Step 3: Combine the information \\
Since we have established that Lana Wilson is a filmmaker and she is from Kirkland, we can conclude:
    Filmmaker(lanaWilson) \( \land \) From(lanaWilson, kirkland)

Step 4: Analyze the conclusion's claim \\
The conclusion states that Miss Americana is not directed by a filmmaker from Kirkland, represented in FOL as:
    \( \neg \) \(\exists t\) x (Filmmaker(x) \( \land \) From(x, kirkland) \( \land \) DirectedBy(missAmericana, x))

However, we have already established that Lana Wilson, who directed Miss Americana, is indeed a filmmaker from Kirkland. This directly contradicts the conclusion.

\textbf{Final Conclusion:} \\
The conclusion "Miss Americana is not directed by a filmmaker from Kirkland" is False.

\textbf{Conditions and prerequisite steps used:}\\
step 1: condition 1, condition 2\\
step 2: condition 5\\
step 3: step 1, step 2\\
step 4: step 3\\

    \color{gray}{/* \textit{Input} */}\\
    ---INPUT---\\
    Question:\\
    \{\textbf{Given\_question}\}\\
    Answer:\\
    \{\textbf{Given\_answer}\}\\
    ---OUTPUT---\\
    \bottomrule
    \end{tabularx}
  \caption{
    The prompt for extracting Conditions and prerequisite steps used in each step of step-by-step solutions.
  }
  \label{tab:100prompt_DAG}
\end{table*}


\end{document}